\documentclass[11pt,a4paper]{article}
\usepackage[hyperref]{emnlp-ijcnlp-2019}
\usepackage{times}
\usepackage{latexsym}

\usepackage{url}

\usepackage{hyperref}
\usepackage{times}
\usepackage{soul}
\usepackage{graphicx}
\usepackage{amsmath}
\usepackage{booktabs}
\usepackage{graphicx}
\usepackage{url}
\usepackage{adjustbox,multirow}
\usepackage{tabularx}
\usepackage{amsmath}
\usepackage{pgf}
\usepackage{color,soul}
\usepackage{float}

\aclfinalcopy

\title{Book Success Prediction with Pretrained Sentence Embeddings and Readability Scores}

\author{Muhammad Khalifa \\
  The University of Michigan \\
  \texttt{khalifam@umich.edu} \\\And
  Aminul Islam \\
  University of Louisiana at Lafayette \\
  \texttt{aminul@louisiana.edu} \\}

\begin{document}

\maketitle

\begin{abstract}
 Predicting the potential success of a book in advance is vital in many applications. This could help both publishers and readers in their decision-making process whether or not a book is worth publishing and reading, respectively. In this paper, we propose a model that leverages pretrained sentence embeddings along with various readability scores for book success prediction. 
  Unlike previous methods, the proposed method requires no count-based, lexical, or syntactic features. Instead, we use a convolutional neural network over pretrained sentence embeddings and leverage different readability scores through a simple concatenation operation. 
  Our proposed model outperforms strong baselines for this task by as large as 6.4\% F1-score points. Moreover, our experiments show that according to our model, only the first 1K sentences are good enough to predict the potential success of books.
  
  \end{abstract}

\section{Introduction}
The ability to predict how likely a book is to succeed is highly valuable for authors, publishers, and readers. For authors, evaluating a book draft before submission to a publisher could save them another rejection letter. For publishers, sifting through all submitted manuscripts is time consuming and there is an obvious need for automating that process. For readers, specially for newly published books, suggestion about whether a book would be interesting or successful is crucial. Moreover, the judgment of editors as to whether to accept a manuscript or not is not always dependable. We know about numerous great writers, such as J.K. Rowling, C.S. Lewis, and Vladimir Nabokov, receiving rejections on books that later turned into worldwide bestsellers. This misjudgment from the publishers' side can greatly be alleviated if we are able to leverage existing book reviews databases through building machine learning and deep learning models that can anticipate how promising a book would be.

Unfortunately, books success prediction is indeed a difficult task. First, many factors determine the success of a book. Some factors come from the book itself such as writing style, clarity, flow and story plot, while other factors are external to the book such as author's portfolio and reputation. Second, from a natural language processing (NLP) perspective, books are typically very long in length compared to other types of documents. For example, an average book could have around 50K words on average. As a result, models that work well for shorter text classification tasks are generally not applicable in this task. Thus, processing a full book in a word-by-word fashion using a Recurrent Neural Network (RNN), for instance, is prohibitive and inefficient \cite{liu2018long}. This is caused by vanishing gradients occurring during training by back-propagation through time (BPTT). One solution is to sample random sentences from the input book and use these sentences as input to the classifier. However, the classifier performance, in this case, can highly vary depending on the random sampling process. Another solution is to divide the input book into chunks of sentences, then aggregate the features within each chunk. In this work, we follow the latter approach.

Previous works on book success prediction have focused on extracting count-based, lexical and syntactic hand-crafted features and used these features for classification~\cite{ashok2013success, maharjan2017multi}. However, the quality of such methods heavily depends on the quality of the features extracted. In addition, while such features may represent the writing style of a given book, they fail to capture semantics, emotions, and plots. The proposed approach in \cite{Maharjan2018LettingEF} focused on modeling the emotion flow throughout the book, arguing that book success relies mainly on the flow of emotions a reader feels while reading. However, emotions are only one aspect of the reading experience and while emotion flow may be an important element in fiction books, it is not the case for non-fiction ones. Our work proposes a method for book success prediction that requires no feature engineering and that takes into account the writing style, content, and semantics.

Our model makes use of transfer learning by applying a pretrained sentence encoder model to embed book sentences. To model book style and readability, we augment the fully-connected layer of a Convolutional Neural Network (CNN) with five different readability scores of the book. We use the dataset published in \cite{maharjan2017multi} and we achieve the state-of-the-art results improving upon the best results published in \cite{Maharjan2018LettingEF}. Our contributions are the followings: 
\begin{itemize}
\item We propose to use CNNs over pretrained sentence embeddings for book success prediction and obtain state-of-the-art performance on the task without any feature engineering.
\item We show that only the first 1K sentences are sufficient to predict the success of a book according to the proposed model.
\item We highlight the connection with the task of book genre identification. We show that sentence embeddings that are good at capturing the separability of book genres display better performance on the book success prediction task. 
\item By augmenting our model with readability scores, we show that readability is a determining factor in book success prediction and that while more readability corresponds to more success, this is not the case for all the readability indices used.
\end{itemize}
The rest of this paper is organized as follows: the related works are discussed in Section~\ref{sec:rw}. Our proposed model is described in Section~\ref{sec:model}. Dataset and experimental results are discussed in Section~\ref{sec:exp}. Direction for future research and conclusion are in Section~\ref{sec:conclusion}.

\section{Related Work}
\label{sec:rw}
Some work has been done on studying writing style and quality. For instance,~\newcite{pitler2008revisiting} studied how various linguistic and syntactic features such as vocabulary and lexical cohesion correlate with text quality and readability.  Furthermore, \newcite{louis2013makes} combined readability, interestingness and content-related features to predict science writing quality. As for book success, in particular, \newcite{ashok2013success} proposed a dataset for book success where book success was determined through the Project Gutenberg\footnote{https://www.gutenberg.org/} download count. They evaluated how different style-related features such as lexical choices, word categories, sentiment, and grammatical rules affect the success of novels. They argued that more success corresponds to less readability, measured in terms of Flesch~\cite{flesch1948new} and Gunning Fog~\cite{gunning1952technique} indices. We obtain similar results with respect to the Automated Readability Index (ARI) and Simple Measure of Gobbledygook (SMOG) index but opposite result on Coleman-Liau Index (CLI).

In~\cite{maharjan2017multi}, the authors released a new dataset for book success prediction based on Goodreads using a more intuitive success measure, that is the Goodreads user rating of the book. They framed book success in a multi-task learning setting by predicting genre and success simultaneously. However, their method was mainly based on hand-crafted syntactic, lexical and count-based features such as Term Frequency - Inverse Document Frequency (TF-IDF) of words and character $n$-grams, and writing density. Another work \cite{Maharjan2018LettingEF} focused more on content rather than style arguing that success is related to emotion variation throughout the book. Thus, a book is encoded as a sequence of emotion aggregated vectors and a Gated Recurrent Unit (GRU) network is employed over the sequence for prediction. However, while they obtained good performance on the Goodreads dataset, emotion variations in a text generally fail to capture writing style or clarity, which are indeed two significant factors in book success.

Transfer learning for NLP tasks, where knowledge from one task is transferred to another task, has produced good results on many NLP tasks~\cite{conneau2017supervised, howard2018universal, devlin2018bert}. BERT~\cite{devlin2018bert} is a bidirectional transformer model~\cite{vaswani2017attention} pretrained on large unlabeled corpus for masked language modeling and next sentence prediction. Through fine-tuning on a target task, BERT gave state-of-the-art performance on many NLP tasks such as sentiment analysis, paraphrase detection, and question answering. BERT was trained on two datasets, namely Wikipedia and the BookCorpus~\cite{moviebook}, which is a collection of around 11K books collected from the web. This makes BERT more suitable for our task. Thus, we employ BERT by fine-tuning it on the Goodreads dataset \cite{maharjan2017multi} and report the results.

\section{Model}
\label{sec:model}
Our proposed model works as follows: Given a book, we use a pretrained sentence encoder to embed each of the book sentences. Then, sentence embeddings are split into near-equal sized chunks. Also, various readability indices are computed on the book content. Then, a Convolutional Neural Network (CNN) is employed for success prediction given both the book embeddings and the readability scores. A complete view of our model is shown in Figure~\ref{Fig:Model}.
\subsection{Universal Sentence Encoder}
Previous work on book success and writing quality prediction has focused extensively on modeling books using hand-crafted style-related features such as word and character $n$-grams~\cite{maharjan2017multi}, or by counting emotionally-expressive words as \cite{Maharjan2018LettingEF}. In this paper, we take an embedding-based approach where we embed book sentences using a pretrained sentence encoder. We use the Universal Sentence Encoder (USE) model proposed in~\cite{cer2018universal}. This encoder consists of a  Deep Averaging Network (DAN)~\cite{iyyer2015deep} where input embeddings for words and bi-grams are first averaged together and then passed through a feed-forward network to produce a 512-dimensional embedding vector. The computation time of DAN is linear in the length of the input sequence, making it suitable for processing long length documents such as books. The USE was trained for various tasks including next sentence prediction, a conversational input-response task and other classification tasks. To assess its performance on our task, we compare the performance of USE to both a bag-of-words baseline and another pretrained sentence embeddings model known as InferSent~\cite{conneau2017supervised}. InferSent was trained on the SNLI dataset~\cite{snli:emnlp2015} to classify pairs of sentences to being either a contradiction, entailment or neutral.

\subsection{CNN} 
Convolutional Neural Networks (CNN) mainly operate by sliding filters on the input representation to compute a set of feature maps. By using multiple filters of different window sizes, CNNs are able to capture various features from the input. Although CNNs were originally used on images, they have shown promising results in various NLP tasks such as Text Classification \cite{kim2014convolutional, gehring2017convolutional}  and other traditional NLP tasks~\cite{collobert2011natural}. 

Since, as we state earlier, book success prediction can be viewed as a text classification task, we conjecture that CNN will be a good fit for the task. We employ a single 1-D convolution layer over the input embeddings to produce a feature map. This feature map is then followed by ReLU non-linearity and max-over-time pooling. ReLU, or Rectified Linear Unit, is an activation function that passes the input directly if it is positive, otherwise, it will output zero.

\begin{figure}[h]
\hspace*{0.5cm}                                         \includegraphics[width=7cm, height=12cm]{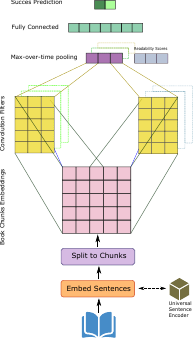}
\caption{The model diagram}
\label{Fig:Model}
\end{figure}

\subsection{Readability}
\cite{louis2013makes} showed that readability features such as token-type ratio and word length significantly impact the writing success. We choose to model readability using five pre-defined readability scores and we incorporate such scores into our neural model by concatenating the resulting 5-dimensional readability vector to the max-over-time pooling layer output and projecting the resulting vector to the classification layer (see Figure~\ref{Fig:Model}). Let $W$ be the number of words in the text, $C$ be the number of characters, $S$ be the number of sentences, $L$ be the number of syllables, and $P$ be the number of polysyllables. Following are the five readability indices we use:

\textbf {Flesch Reading Ease Score} \cite{flesch1948new} which is a number in the range 1-100 that estimates roughly what level of education someone will need to be able to easily read a piece of text. It is computed as a function of the number of words, sentences, and syllables. Higher score indicates easier material.
\begin{equation}
\begin{split}
\text{FRES} = 206.835 - 
1.015 \times \frac{W}{S} -  
84.6 \times \frac{L}{W}
\end{split} \label{eq1}
\end{equation}

\textbf {Flesch Kincaid Grade}  \cite{kincaid1975derivation} This grade is more common in the field of education. Unlike FRES, higher scores correspond to higher difficulty.
\begin{equation}
\begin{split}
\text{FKG} = 0.39 \times \frac{W}{S} -  
11.8 \times \frac{L}{W} - 15.59
\end{split} \label{eq2}
\end{equation}

\textbf {SMOG} 
\cite{mc1969smog} which is a measure of readability estimating the years of education needed to understand a piece of writing.
\begin{equation}
\text{SMOG}= 1.0430 \times \sqrt[]{P\frac{30}{S}} + 3.1291 \label{eq3}
\end{equation}

\textbf {Coleman-Liau Index}
\cite{coleman1975computer} It is similar to Flesch Kincaid in that it predicts the U.S. grade level required to understand the text.
\begin{equation}
\text{CLI}= 0.0588C' - 0.296S' - 15.8 \label{eq4}
\end{equation}
Where $C'$ is the average number of characters per 100 words and $S'$ is the average number of sentences per 100 words. 

\textbf {Automated Readability Index}
\cite{senter1967automated}
is an index similar to Coleman-Liau and Flesch-Kincaid.

\begin{equation}
\begin{split}
\text{ARI}= 4.71 \times \frac{C}{W} 
+ 0.5 \times \frac{W}{S} - 21.43
\end{split} \label{eq5}
\end{equation}

\section{Experiments}
\label{sec:exp}
In this section, we first describe the dataset we use for training and evaluation of our model. Then, we describe the baseline and other state-of-the-art models that we compare our model to. Finally, we show the results obtained using the test dataset and discuss various interpretations of these results.
\subsection{Data}
There are two publicly available datasets for books success prediction, namely EMNLP13 \cite{ashok2013success} and Goodreads \cite{maharjan2017multi}. The EMNLP13 dataset contains 800 books, while the Goodreads dataset contain 1,003. The main difference between the two datasets is in the definition of success. The success prediction ground-truth of EMNLP13 was based on the book download count on Project Gutenberg, while that of the Goodreads dataset was based on the Goodreads user rating. In \cite{maharjan2017multi}, it was argued that download counts are not a solid indicator of book success; the authors found 142 books with different success labels in each dataset, 19.7\% of which have more than 100 reviews. We choose to train and test our model using the Goodreads dataset for two reasons. First, it is larger than the EMNLP13 dataset (around 200 more books). This makes it more suitable for neural models such as ours, which require more data. Second, the download count used in the EMNLP13 dataset measures only one aspect of success, that is popularity. Other aspects such as the effect of the book on the reader can not be determined solely based on the download count. Therefore, we choose the Goodreads dataset since we believe the Goodreads rating can be a more comprehensive measure that can capture popularity (the number of voters for a given book is taken into account in the Goodreads rating) in addition to other aspects. The books in the Goodreads dataset are from eight different genres and have been rated by at least 10 people. A book is labeled successful if its average Goodreads rating is 3.5 or more (The Goodreads rating scale is 1-5). Otherwise, it is labeled as unsuccessful. Table~\ref{dataset-table} shows the Goodreads dataset statistics. As shown in the table,  the positive (successful) class count is almost double than that of the negative (unsuccessful) class count.
\renewcommand{\tabcolsep}{3pt}
\begin{table}
\noindent
\begin{tabular}{|l|l|l|l|}
 \hline
 \bf Genre & \bf Unsuccessful & \bf Successful & \bf Total \\ \hline
Detec. Mystery & 60 & 46 & 106  \\
Drama & 29 & 70 & 99 \\
Fiction & 30 & 81 & 111 \\
Hist. Fiction &  16 &  65 &  81  \\
Love Stories &  20 & 60 &  80  \\
Poetry & 23 &  158 & 181 \\
Sci. Fiction &  48 & 39  &  87  \\
Short Stories &  123 &  135 & 258  \\
\hline
Total &  349 & 654  &  1,003  \\
\hline
\end{tabular}
\caption{\label{dataset-table} Goodreads dataset statistics \cite{maharjan2017multi}.}
\end{table}

\subsection{Baseline Models}
We compare our approach to several baseline models:

\textbf{Majority Class:} Predicting the more frequent class (successful) for all the books.

\textbf{Book2Vec:} Sentence embeddings are averaged to obtain a single vector which is then fed to a 2-layer feed-forward network for prediction.

\textbf{Bi-LSTM:} The USE chunk embeddings sequence is processed with a one-layer Bi-LSTM with attention on the hidden states similar to \cite{lin2017structured}.

\subsection{Competing Methods}
We compare our approach to the following two state-of-the-art models:

\textbf{ST-HF:} The best single-task model proposed by \cite{maharjan2017multi}, which employs various types of hand-crafted features including sentiment, sensitivity, attention, pleasantness, aptitude, polarity, and writing density.

\textbf{Emotion Flow:} This model is proposed by \cite{Maharjan2018LettingEF} and is comprised of a bidirectional GRU with attention similar to \cite{lin2017structured}.

\textbf{BERT:} We fine-tune the BERT uncased base model (11 layers, total parameters = 110M) \cite{devlin2018bert} on our task. Since BERT is limited to a maximum sequence length of 512 tokens, we split each book into 50 chunks of almost equal size, then we randomly sample a sentence from each chunk to obtain 50 sentences. This is done to ensure that the sampled sentences span the whole book. Then the 50 sampled sentences are concatenated and fed to BERT. We fine-tuned BERT for 150 epochs using Adam optimizer with a learning rate of 0.00001.

\subsection{Experimental Setup}
Each book is partitioned to 50 chunks where each chunk is a collection of sentences. We experiment both on the first 1K sentences and full book as in \cite{Maharjan2018LettingEF}.

To compute the readability scores, we use \emph{TextStat}\footnote{https://github.com/shivam5992/textstat}, which is a python library that can compute various statistics from text. We compute the sentence embeddings using the pretrained Universal Sentence Encoder model available on 
Tensorflow Hub\footnote{https://tfhub.dev/google/universal-sentence-encoder/1}
to encode each sentence into a 512-dimensional vector. After computing embeddings for individual sentences, we average the sentence embeddings within each chunk to obtain a final vector for that chunk. Thus, each book is modeled as a sequence of chunk embedding vectors.

We randomly sample a 20\% of the training dataset to obtain a validation set. For the CNN model, we use the validation set to obtain the best set of possible hyper-parameters. We found that using 20 filters of sizes 2, 3, 5, and 7 and concatenating their max-over-time pooling output gives the best results. We use 50 units as the size of the fully connected layer. We also use a Dropout \cite{srivastava2014dropout} with probability 0.6 over the convolution filters. We optimize using Adam \cite{Kingma2014Adam:Optimization} with a learning rate of 0.0009 and $\beta_1$ and $\beta_2$ of 0.9 and 0.999, respectively. During training, we keep track of the best model on the validation set and use it on the test set.

\begin{figure*}[htb!]
\hspace*{-2.8cm}  
\centering
\includegraphics[width=21cm, height=6cm]{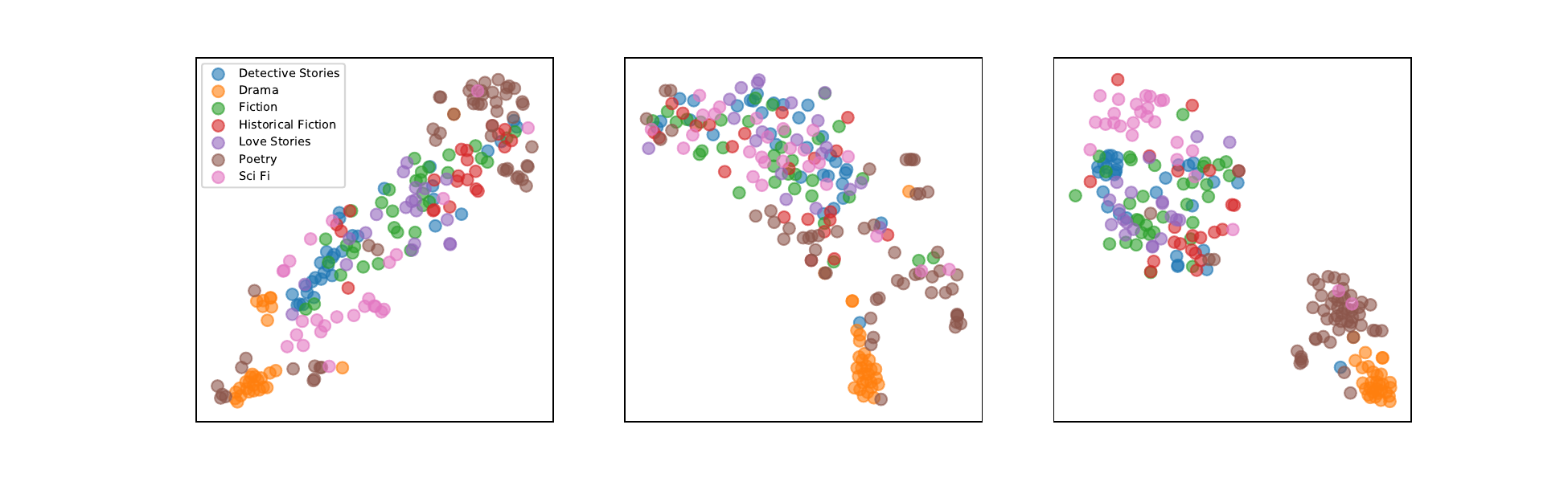}
\caption{\label{fig:t-sne} t-SNE Plot of the average book embeddings on the test set. \textbf{Right} is the USE embeddings. \textbf{Middle} is bag-of-words embeddings. \textbf{Left} is the InferSent embeddings.}
\end{figure*}

\subsection{Results and Discussion}

\subsubsection{Success Prediction}
Table~\ref{Tab:1k} shows the results of our models against the two state-of-the-art models and three baseline models using the first 1K sentences. To measure the model accuracy, we use the weighted F1-score (the harmonic mean of the precision and the recall) where each class score is weighted by the class count. Clearly, the CNN model without readability scores outperforms other baselines with a weighted F1 score of 0.674. When book readability scores are included with the CNN model, the weighted F1 score increases to 0.720 (which is comparable to the ST-HF model in \cite{maharjan2017multi}) giving the best performance.

\begin{table}[h]
\begin{center}
\begin{tabular}{|l|l|l|}
\hline
\bf Model & \bf F1  \\ \hline
Majority Class  & 0.506 \\
Book2Vec  & 0.635 \\ 
Bi-LSTM   & 0.659 \\
Emotion Flow
\cite{Maharjan2018LettingEF} &  0.656 \\
BERT  &  0.660 \\
\hline
CNN (ours)  &  0.674  \\
CNN with Readability (ours) & \bf 0.720* \\

\hline
\end{tabular}
\end{center}
\caption{\label{Tab:1k} Weighted F1-score on the test set using only the first 1K sentences. *McNemar Significance test between Book2Vec and this model with $p < 0.05$.}
\end{table}

\begin{table}[h]
\begin{center}
\begin{tabular}{|l|l|l|}
\hline
\bf Genre & \bf F1  \\
\hline
Detective Mystery & 0.597 \\
Drama & 0.795 \\
Fiction & 0.671 \\
Historical Fiction & 0.767  \\
Love Stories & 0.736  \\
Poetry & 0.795 \\
Science Fiction &  0.366  \\
Short Stories & 0.745  \\
\hline
\end{tabular}
\end{center}
\caption{\label{Tab:1k-genre}
 Weighted F1-score on test set for each genre using our model trained on first 1K sentences.}
\end{table}

Table~\ref{Tab:1k-genre} Shows our best model performance per genre. Our model seems to work best on Poetry, Love Stories, and Drama books. Intuitively, these three genres typically exhibit high emotional content and variations which are well-captured by the CNN filters. On the other hand, the success of genres such as Mystery and Science Fiction is typically based on the story plot and twists, requiring a much greater understanding of the content than what is captured by our model.

\begin{table}[h]
\begin{center}
\begin{tabular}{|l|l|}
\hline
\bf Model &  \bf F1  \\ \hline
Majority Class (Baseline) & 0.506 \\
Book2Vec (Baseline) & 0.649 \\ 
Bi-LSTM (Baseline) & 0.676  \\
ST-HF \cite{maharjan2017multi} & \textbf{0.720} \\
Emotion Flow \cite{Maharjan2018LettingEF} &   0.690 \\
BERT &   0.654 \\
\hline
CNN &  0.685  \\
CNN with Readability & \bf 0.708 \\

\hline
\end{tabular}
\end{center}
\caption{\label{Tab:fullbook} Weighted F1-score on the test set using the full book.}
\end{table}

\begin{table}[h]
\begin{center}
\begin{tabular}{|l|l|}
\hline
\bf Section & \bf F1  \\ \hline
First 1K sentences & \bf 0.720 \\
First 5K sentences & 0.685 \\ 
First 10K sentences  & 0.698  \\ 
Last 1K sentences & 0.672 \\
Full Book & 0.708 \\

\hline
\end{tabular}
\end{center}
\caption{\label{Tab:sentences} Weighted F1 score on the test set obtained when using different sections from each book as input.}
\end{table}

\begin{figure}[htb!]
\centering
\scalebox{0.57}{\input{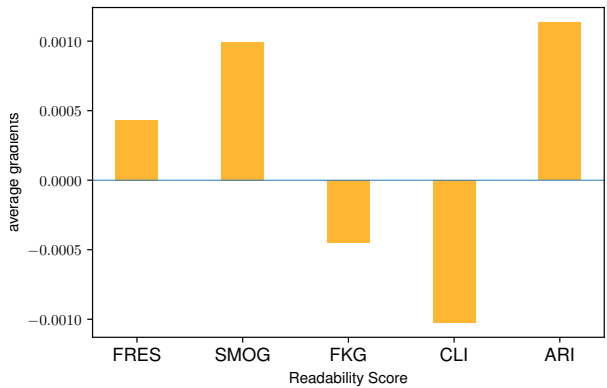}}
\caption{Contribution of the five readability scores to success prediction.}
\label{Fig:Read}
\end{figure}

We further compare our model against the other models while using the whole book in Table~\ref{Tab:fullbook}. Clearly, our model outperforms the other two but underperforms ST-HF. Interestingly, training the model with smaller portion of the book (such as first 1K sentences) gives better performance over using the full book. We conjecture that this is due to the fact that, in the full-book case, averaging the embeddings of larger number of sentences within a chunk tends to weaken the contribution of each sentence within that chunk leading to loss of information.

We further study book success prediction using different number of sentences from different location within a book. We conduct further experiments by training our best model on the first 5K, 10K and the last 1K sentences. Table~\ref{Tab:sentences} shows the results. We notice that using the first 1K sentences only performs better than using the first 5K and 10K sentences and, more interestingly, the last 1K sentences. This could point out to the conclusion that book openings are a better indicator of success than book endings.

\subsubsection{Sentence Embeddings}
To evaluate USE embeddings, we compare a bag-of-words model based on GloVe word embeddings~\cite{pennington2014glove}, Infersent~\cite{conneau2017supervised}, and USE embeddings on the book success task for the first 1K sentences shown in Table~\ref{Tab:1k-emb}.
\begin{table}[h]
\begin{center}
\begin{tabular}{|l|l|l|}
\hline \bf Model & \bf Val  F1 & \bf Test F1  \\ \hline
BOW & 0.720 & 0.640  \\
InferSent & 0.743 & 0.667  \\
USE & \bf 0.817 & \bf 0.674  \\
\hline
\end{tabular}
\end{center}
\caption{\label{Tab:1k-emb}
 Comparison of Different Sentence Embedding Models on the CNN model.}
\end{table}

We also show a t-SNE plot of the averaged embeddings plotting according to genres in Figure~\ref{fig:t-sne}. Clearly, the genre differences are reflected in USE embeddings (Right) showing that these embeddings are more able to capture the content variation across different genres than the other two embeddings. We have two observations based on the test set: First, USE embeddings give best performance for book success prediction. Second, USE embeddings model the genre distribution of books the best. This could be an indicator of a strong connection between the two tasks and is supported by the results in \cite{maharjan2017multi} and \cite{Maharjan2018LettingEF}, where using book genre identification as an auxiliary task to book success prediction helped improve the prediction accuracy.

\subsubsection{Readability Indices}

To measure the contribution of readability indices to success prediction, we compute the gradients of the success variable in the output layer with respect to each readability index on the test set. Figure~\ref{Fig:Read} shows the average of gradients computed for each readability index. We can see positive gradients for SMOG, ARI, and FRES but negative gradients for FKG and CLI. As shown, the SMOG, CLI, and ARI have the largest gradients compared to the others. Interestingly, while low value of CLI and FKG (i.e., more readable) indicates more success, high value of ARI and SMOG (i.e., less readable) also indicates more success. Obviously, high value of FRES (i.e., more readable) indicates more success. This poses an important question: do these opposing indices measure different aspects of readability such that one aspect is positively correlated with writing success while the other is negatively correlated? Looking at the Equations~\ref{eq4} and \ref{eq5} for computing CLI and ARI (which have opposite gradient directions), we find out that they differ with respect to the relationship between words and sentences. While ARI uses the average number of words per sentences, the CLI uses the conjugate, that is, the average number of sentences per $n$ words. Interestingly, this observation can be interpreted in a way such that more successful books tend to have large number of words per sentences but small number of sentences per $n$ words.

\section{Conclusion}
\label{sec:conclusion}
In this paper, we proposed to use a Convolutional Neural Network and readability scores for book success prediction. The CNN is used to process sentence embeddings obtained from the pretrained Universal Sentence Encoder. Our method outperforms strong baseline methods without using any feature engineering and performs comparably well to the state-of-the-art methods. In addition, our results show that the success prediction accuracy using a portion of the book is better than it is when using the whole book.

Moreover, by visualizing the book embeddings based on genre, we argued that embeddings that better separate books based on genre gave better results on book success prediction than other embeddings. We also showed that while more readability corresponds to more success according to some readability indices such as Coleman-Liau Index (CLI) and Flesch Kincaid Grade (FKG), this was not the case for other indices such as Automated Readability Index (ARI) and Simple Measure of Gobbledygook (SMOG) index. By taking CLI and ARI as two examples, we argued that it is better for a book to have high words-per-sentences ratio and low sentences-per-words ratio. 

As future work, employing more pretrained language models for sentence embedding, such as BERT and GPT2, is worthy of exploring and would likely give better results. Another follow up work is to investigate the connection between readability and success with a more detailed empirical analysis.

\bibliography{ref}
\bibliographystyle{acl_natbib}

\end{document}